\documentclass[10pt,conference,final]{IEEEtran}
\usepackage{cite}
\usepackage{amsmath,amssymb,amsfonts}
\usepackage{algorithmic}
\usepackage{graphicx}
\usepackage{textcomp}
\usepackage{xcolor}
\usepackage{url}
\usepackage{hyperref}
\usepackage{fixme}

\fxsetup{ status=draft, theme=color, inline, margin=False}
\FXRegisterAuthor{shang}{ashang}{\color{red}S}
\FXRegisterAuthor{polo}{apolo}{\color{red}P}
\FXRegisterAuthor{cory}{acory}{\color{red}C}
\FXRegisterAuthor{jason}{ajason}{\color{red}J}
\begin{document}
%\title{Realistic Evaluation of Evasion Attacks in CARLA \\ {\Large Talk Proposal}}
\title{Talk Proposal: Towards the Realistic Evaluation of Evasion Attacks using CARLA}
\author{
\IEEEauthorblockN{
    Cory Cornelius\IEEEauthorrefmark{1}, Jason Martin\IEEEauthorrefmark{2}
}
\IEEEauthorblockA{
    Intel Corporation \\
    Hillsboro, OR, USA \\
    \IEEEauthorrefmark{1}\href{mailto:cory.cornelius@intel.com}{cory.cornelius@intel.com}, \IEEEauthorrefmark{2}\href{mailto:jason.martin@intel.com}{jason.martin@intel.com}, 
}
\and
\IEEEauthorblockN{
    Shang-Tse Chen\IEEEauthorrefmark{3}, Duen Horng (Polo) Chau\IEEEauthorrefmark{4}
}
\IEEEauthorblockA{
    Georgia Institute of Technology \\
    Atlanta, GA, USA \\
    \IEEEauthorrefmark{3}\href{mailto:schen351@gatech.edu}{schen351@gatech.edu}, \IEEEauthorrefmark{4}\href{mailto:polo@gatech.edu}{polo@gatech.edu}}
}

\maketitle

\begin{abstract}
In this talk we describe our content-preserving attack on object detectors, ShapeShifter~\cite{chen2018shapeshifter}, and demonstrate how to evaluate this threat in realistic scenarios.
We describe how we use CARLA~\cite{dosovitskiy2017carla}, a realistic urban driving simulator, to create these scenarios, and how we use ShapeShifter to generate content-preserving attacks against those scenarios.
\end{abstract}


\section{Introduction}
Most adversarial machine learning research focuses on indistinguishable perturbations 
while few focus on more realistic threats like content-preserving perturbations and non-suspicious inputs~\cite{gilmer2018rules}.
%while few focus on more realistic threats like content-preserving perturbations, non-suspicious inputs, content-constrained inputs, and unconstrained inputs~\cite{gilmer2018rules}.
Although some research does focus on these more realistic threats~\cite{zeng2017beyond}, they are often limited to the classification task even though many safety- and security-critical systems need to localize, in addition to classify, objects.
These systems often perceive the physical world with sensors, and most attacks do not account for the characteristics of this sensing pipeline and thus fail to remain adversarial when physically realized.
%These systems often perceive the physical world using sensors, and most attacks fail to account for the characteristics of this sensing pipeline and the idiosyncrasies of the sensed environment.
%When physically realized, many attacks fail to remain adversarial.

\section{Our ShapeShifter Attack}
\begin{figure}[ht]
    \centering
    \includegraphics[width=0.49\textwidth]{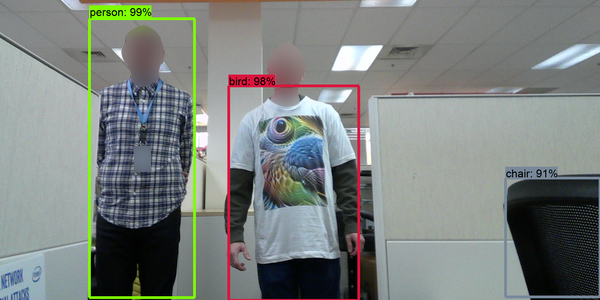}
    \caption{An accessorized~\cite{sharif2016accessorize} t-shirt created by our ShapeShifter attack~\cite{chen2018shapeshifter} causes the object detector to see a bird rather than a person. The person on the left is correctly detected with 99\% confidence, while the on the right with the accessorized t-shirt on is mis-detected as a \textit{bird} with 98\% confidence. We set the detection threshold at 50\%. Faces anonymized.}
    %\caption{Our ShapeShifter attack~\cite{chen2018shapeshifter} generates non-suspicious content-preserving perturbations that cause the object detector to see a bird rather than a person.}
    \label{fig:shapeshifter}
\end{figure}

Our recent work, ShapeShifter~\cite{chen2018shapeshifter}, creates content-preserving physically-realizable adversarial stop signs that are mis-detected as other objects by the Faster R-CNN Inception-v2 object detector trained on the MS-COCO dataset.
%Our recent work, ShapeShifter~\cite{chen2018shapeshifter}, shows how to attack object detectors with content-preserving physically-realizable perturbations.
%We used our attack to create perturbed stop signs that cause an object detector to ``see'' another object than the stop sign.
%By fabricating these stop signs and recording them with a camera, we showed our attack remains adversarial when physically realized.
More recently, we showed how to make this attack an accessory~\cite{sharif2016accessorize} by limiting the perturbation to the printable area of a t-shirt as shown in Figure~\ref{fig:shapeshifter}.

To create these attacks, we first had to digitally craft them, then physically fabricate them, and finally test them in the real world.
This process is straightforward assuming one knows how to accurately model these real-world transformations that Expectation over Transformation~\cite{athalye2017synthesizing} requires, a method ShapeShifter relies upon.
More often there was a non-trivial amount of iteration between crafting, fabricating, and testing.
Additionally, choosing how to perturb objects that are realistically and physically constrained (e.g., attackers cannot typically perturb the sky or even all parts of the object in non-suspicious ways) took some consideration.
Our experience has led us to seek means to reduce these difficulties.

\section{The CARLA Simulator}
\begin{figure}[ht]
    \centering
    \includegraphics[width=0.49\textwidth]{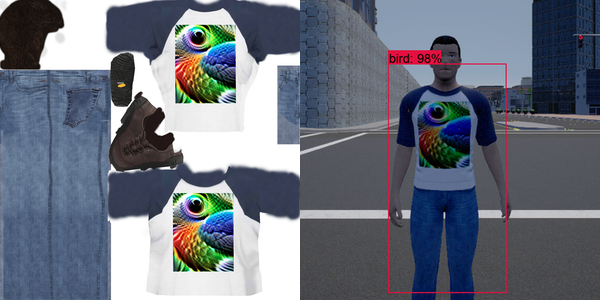}
    \caption{We transferred the attack onto the CARLA pedestrian model. The perturbation remained adversarial for a limited set of camera positions and orientations. The pedestrian model is incorrectly detected as a \textit{bird} with 98\% confidence.}
    %\caption{We transferred this attack on the CARLA pedestrian model by manually adding the perturbation onto the model's texture. We found the perturbation remained adversarial for a limited set of camera positions and orientations.}
    \label{fig:shapeshifter-transfer}
\end{figure}

\begin{figure}[ht]
    \centering
    \includegraphics[width=0.49\textwidth]{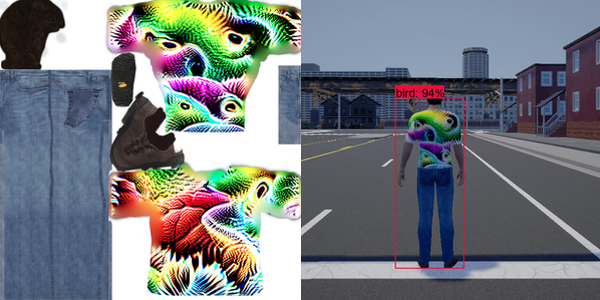}
    \caption{CARLA enables us to perturb the environment with realistic constraints. We created full t-shirt perturbation that is adversarial from different camera positions and orientations. The pedestrian model is mis-detected as a \textit{bird} with 94\% confidence at a further distance than the transferred perturbation in Figure~\ref{fig:shapeshifter-transfer}.}
    \label{fig:shapeshifter-carla}
\end{figure}

CARLA is an open-source realistic driving simulator.
% built upon the Unreal Engine. 
It provides a convenient way to train autonomous driving agents with a variety of different sensor suites (cameras, LIDAR, depth, etc.), and test them in realistic traffic scenarios.
%It provides all of the code and assets to effectively train autonomous driving agents using a variety of different sensor suites (cameras, LIDAR, depth, etc.), and to test them in realistic traffic scenarios.
We can reduce the time between digitally crafting a perturbation and testing it against real scenarios by using CARLA as a simulator of the real-world.
In some cases, we have found our attacks already work in CARLA as shown in Figure~\ref{fig:shapeshifter-transfer}.
CARLA has also enabled us to craft new perturbations that were previously difficult to model as shown in Figure~\ref{fig:shapeshifter-carla}.
The reproducibility of the CARLA scenarios enables us to better understand the efficacy of our attacks across a variety of environmental conditions that CARLA already supports.

Our purpose in creating these attacks and evaluating them under realistic scenarios is to understand whether adversarial examples pose a real threat to real systems.
While it is true that current adversaries might use easier means to subvert these systems (e.g., by knocking over a stop sign), we should not underestimate what methods attackers will undertake to achieve their means.

\begin{figure}[ht]
    \centering
    \includegraphics[width=0.49\textwidth]{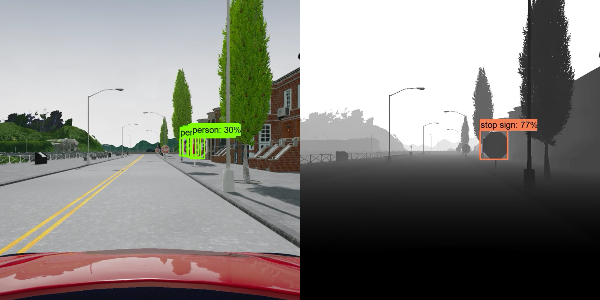}
    \caption{In autonomous driving systems, there is reason to believe adversarial examples like ours can be mitigated using multi-modal sensing. Here we show an off-the-shelf object detector detecteing a \textit{stop sign} using only the depth channel. CARLA enables us to quickly test and prototype these kind of defenses.}
    \label{fig:shapeshifter-depth}
\end{figure}

We also believe the creation of these realistic evaluations will increase our defensive capabilities.
To date it is unclear whether the most successful defense, adversarial training~\cite{madry2017towards}, can even scale to ShapeShifter-style perturbations.
A more recent defense has shown some defensive capability against ShapeShifter-like attacks on image classifiers~\cite{chou2018sentinet}.
However, neither of these defenses think about the system as a whole.
By creating these realistic evaluations in CARLA, defenders can think more holistically about potential strategies.
%By creating these realistic evaluations in CARLA, we, now as defenders, can think more holistically about potential defenses.
For example, CARLA enables us to quickly experiment with different sensing modes.
%For example, CARLA enables us to toggle between different sensing modes and quickly experiment with them.
We have found that the depth channel might provide some defensive capability against pixel perturbations when the object of interest has a distinct shape as shown in Figure~\ref{fig:shapeshifter-depth}.
%Figure~\ref{fig:shapeshifter-depth} shows what an object detector sees from the depth channel alone.
From a system-level perspective, perhaps this is an acceptable defense, because now the attacker must now perturb the geometry of the object in addition to its color.

\section*{Acknowledgements}
This work was supported by gift from Intel, for Intel Science \& Technology Center for Adversary-Resilient Security Analytics (ISTC-ARSA) at Georgia Tech.

\bibliographystyle{hieeetr}
\bibliography{bibliography}

\vspace*{\fill}
\newpage
\vspace*{\fill}

\end{document}